\documentclass[journal]{IEEEtran}
\usepackage{microtype}                 % use micro-typography (slightly more compact, better to read)
\PassOptionsToPackage{warn}{textcomp}  % to address font issues with \textrightarrow
\usepackage{tikz}
\usepackage{comment}
\usepackage{amsmath,amssymb} % define this before the line numbering.
\usepackage{color}
\usepackage[switch]{lineno}
\usepackage{comment}
\usepackage{bbm}
\usepackage{amsmath, xparse}
\usepackage{here}
\usepackage{xcolor}
\usepackage{subfig}
\usepackage{bm}
\usepackage{cite}
\usepackage{multirow}
\usepackage{ctable} 
\usepackage{arydshln}
\usepackage{algorithm}
\usepackage{algorithmic}
\usepackage{enumitem}
\newcommand{\guMR}[1]{{\textcolor{black}{#1}}}
\newcommand{\guPr}[1]{{\textcolor{black}{#1}}}
\newcommand\norm[1]{\left\lVert#1\right\rVert}

\ifCLASSINFOpdf
\else
\fi
\begin{document}
\title{Orientation-Aware Leg Movement Learning for Action-Driven Human Motion Prediction}

%Exploring Realistic Leg Dynamics for Orientation-Aware Action Transition Learning

\author{Chunzhi~Gu,~\IEEEmembership{Member,~IEEE,}
        Chao~Zhang,~\IEEEmembership{Member,~IEEE,}% <-this % stops a space
        ~and~Shigeru~Kuriyama,~\IEEEmembership{Member,~IEEE}% <-this % stops a space
\thanks{C. Gu* and S. Kuriyama are with the Department of Computer Science and Engineering, Toyohashi University of Technology, Toyohashi, Japan (e-mails: gu@cs.tut.ac.jp, sk@tut.jp).}% <-this % stops a space
\thanks{C. Zhang is with the School of Engineering, University of Fukui, Fukui, Japan (zhang@u-fukui.ac.jp).}}
%\thanks{Manuscript received April xx, xx; revised August xx, xx, (Corresponding author: C. Gu.)}}


\markboth{}%
{Shell \MakeLowercase{\textit{et al.}}: Bare Demo of IEEEtran.cls for IEEE Journals}



\maketitle

\begin{abstract}
The task of action-driven human motion prediction aims to forecast future human motion \guPr{based on} the observed sequence while respecting the given action label. It requires modeling not only the stochasticity within human motion but the smooth yet realistic transition between multiple action labels. However, the fact that most datasets do not contain such transition data complicates this task. Existing work tackles this issue by learning a smoothness prior to simply promote smooth transitions, yet doing so can result in unnatural transitions especially when the history and predicted motions differ significantly in orientations. In this paper, we argue that valid human motion transitions should incorporate realistic leg movements to handle orientation changes, and cast it as an action-conditioned in-betweening (ACB) learning task to encourage transition naturalness. Because modeling all possible transitions is virtually unreasonable, our ACB is only performed on very few selected action classes with active gait motions, such as ``\textit{Walk}'' or ``\textit{Run}''. Specifically, we follow a two-stage forecasting strategy by first employing the motion diffusion model to generate the target motion with a specified future action, and then producing the in-betweening to smoothly connect the observation and prediction to eventually address motion prediction. Our method is completely free from the labeled motion transition data during training. To show the robustness of our approach, we generalize our trained in-betweening learning model on one dataset to two unseen large-scale motion datasets to produce natural transitions. Extensive \guPr{experimental evaluations} on three benchmark datasets demonstrate that our method yields the state-of-the-art performance in terms of visual quality, prediction accuracy, and action faithfulness.

%We formulate a VAE-based in-betweening learning framework 

\end{abstract}


\IEEEpeerreviewmaketitle

\section{Introduction}
\IEEEPARstart{H}{uman} motion modeling is a fundamental task in computer vision and graphics and has been extensively studied due to its wide real-world applications. In particular, understanding and modeling human behavior facilitates applications in \guPr{the} robotics \cite{zhang2022reinforcement}, sports \cite{zhang2019predicting}, \guPr{user interface} \cite{wu2020futurepong}, and animation industries \cite{min2009interactive}. Among various modeling tasks, human motion prediction has received active attention. It aims to forecast future human motion based on the given history observation sequence. A powerful human motion prediction model has a profound impact on safety-practical applications, such as autonomous driving. 

Since human motion naturally involves different action types, recent techniques have advanced towards predicting action-conditioned human motion to better promote the progress of autonomous systems. Considering the stochasticity in human motion, this task is introduced as action-driven stochastic human motion prediction, with the goal of characterizing multiple future motions with specified action \guPr{labels} based on the \guPr{historical} observation. Different from the action-agnostic prediction scenario, action-driven prediction involves modeling the transition between multiple action labels. However, most human motion datasets do not include cross-label action transition and consist solely of single action clips. Even though the recently released dataset BABEL \cite{punnakkal2021babel}  partially contains such transition data, the covered transition types are highly limited, which, as reported in \cite{mao2022weakly}, results in poor generality. 

\begin{figure}[t]
\begin{center}
\includegraphics[width=1.0\linewidth]{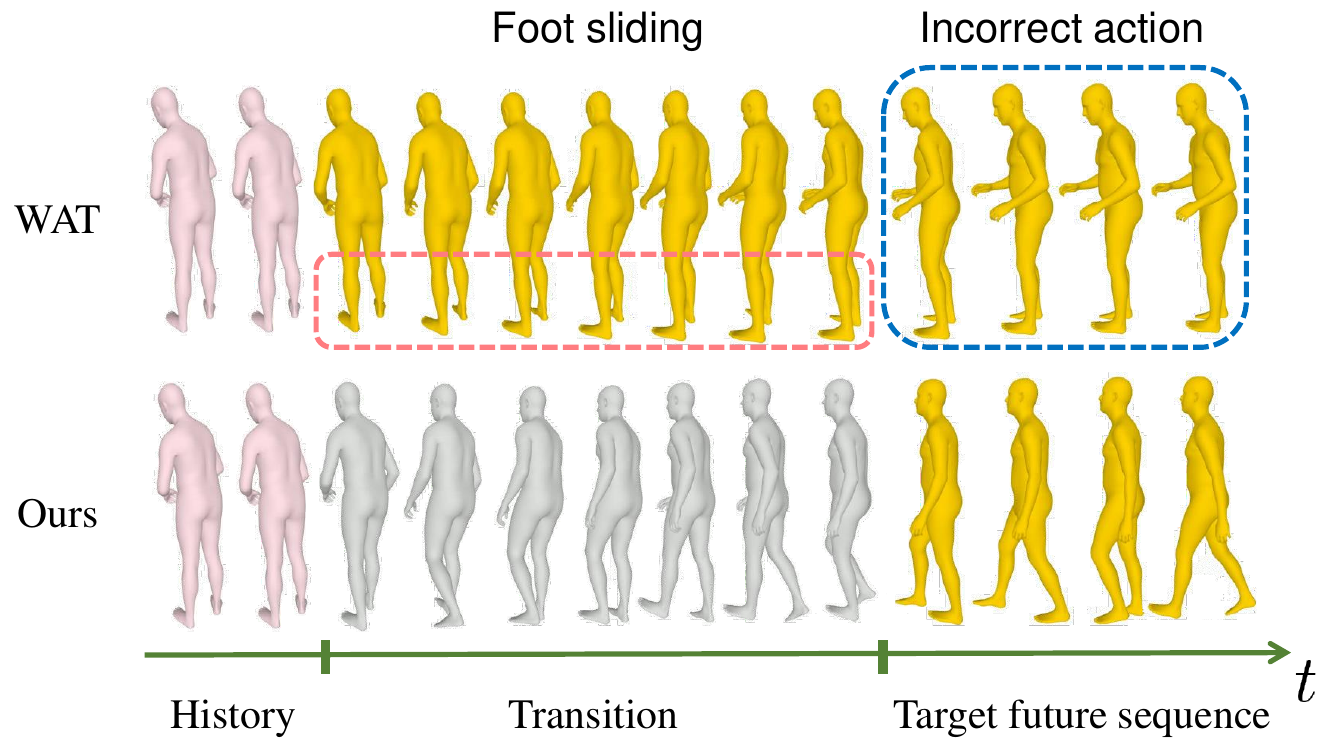}
\end{center}
\caption{An example of \guPr{the} action-driven prediction performance of \guPr{our method compared} against WAT \cite{mao2022weakly}. Both methods predict the future sequence given the action  ``\textit{Walk}''. WAT \cite{mao2022weakly} suffers from unnatural transition (e.g., foot sliding) or poor label faithfulness. Our method jointly yields natural transition with plausible leg movements and respects the action label better.}
\label{fig:introduction}
\end{figure}

As expecting the dataset to \guPr{contain} all possible action transitions is less feasible, the pioneering work \cite{mao2022weakly} proposes modeling the transition by simply encouraging the smoothness. It first prepares the training data (i.e., history and future motions) by randomly pairing the history sequence with another future sequence whose action label is different. \guPr{It then} leverages the discrete cosine transform (DCT) to construct a smoothness prior as weak supervision to promote transition smoothness between inconsistent history-future motion pairs. Despite the conceptual feasibility, it can quickly lead to the generation of unnatural motion specifically when the orientations of the history and future motions differ significantly, as presented in Fig. \ref{fig:introduction}(top row). 

In this paper, we propose a novel strategy by casting the transition learning as an action-conditioned in-betweening (ACB) learning problem to address orientation misalignment of transition. It is inspired by the fact that human movements naturally involve active leg dynamics to bridge the gap between large orientation changes. We thus develop a Variantional Auto-Encoder (VAE) in-betweening learning framework that aims to smoothly connect the given start and end human motions with natural leg movements. Since the transition among actions is inherently stochastic, we further condition our VAE on different in-betweening labels to allow for versatile transition types, and dub our model as Action-conditioned in-Betweening VAE (AinB-VAE). Specifically, AinB-VAE is trained on a limited set of motion clips \guPr{that} involve action types with rich leg dynamics capable of handling orientation changes, \guPr{such as} \textit{Run} and \textit{Walk}, to effectively leverage existing datasets. To further utilize orientation information, we propose an orientation-warping module for AinB-VAE to show awareness of large orientation bias during decoding. In summary, our core idea is to learn motion in-betweening on individual motion clips, which are broadly covered in existing datasets, to eventually address the lack of annotated transition data during training.

Since our goal is to provide a better solution to action-driven stochastic motion prediction, we follow the two-stage forecasting policy by separating the generation of future motion and transition. We first synthesize a sequence with \guPr{a} specified future action label, and then use the trained AinB-VAE to connect the given history and the produced motions to fulfill the task of prediction. As depicted in Fig. \ref{fig:introduction}(2nd row), this allows for realistic transition among different actions. In particular, we utilize the Motion Diffusion Model \cite{tevet2023human} for high-fidelity human motion generation. Unlike \cite{mao2022weakly}, our method enables diversity in terms of both transition and prediction. The transition further includes inter- and intra-class diversity. To improve the diversity of intra-class transition, we additionally adapt to our task an in-betweening sampler that maps the conditions to diversified in-betweenings with the same transition action label.

We evaluate our method on three large-scale human motion datasets: BABEL \cite{punnakkal2021babel}, HumanAct12 \cite{guo2020action2motion}, and NTU RGB-D \cite{liu2019ntu}, all of which include \guPr{per-sequence} action label. Since HumanAct12 and NTU RGB-D contain very limited or no motion data with sufficient gait movements, we use our trained AinB-VAE on BABEL on the other two datasets for transition generation to show the generalization capability. Experimental results show that our model achieves state-of-the-art performance on the task of action-driven human motion prediction in terms of perceptual similarity, prediction accuracy, and action faithfulness. 

Our contributions can be summarized as follows: (1) We propose 
casting the action transition learning as an action conditioned in-betweening learning task to allow generating plausible gait transition without demanding the annotated transition data; (2) We propose a VAE-based in-betweening learning framework with an orientation-aware decoding module to pursue realistic leg movements\guPr{,} even for large changes of body orientation; (3) We present a two-stage pipeline for action-driven human motion prediction that contributes to natural yet diverse generation for both motion transition and prediction; (4) We report extensive experimental results to show the effectiveness of our method qualitatively and quantitatively in both motion in-betweening and prediction tasks. 

%We argue that the change of orientation in human motion should naturally involve the required leg dynamics to form realistic transition and propose a VAE-based framework to specifically learn in-betweenings from the actions with frequent leg movements;

\section{Related Work}
In this section, we first review previous human motion prediction techniques. We then discuss some literature on human motion in-betweening. Finally, we review some transition learning approaches for human motion synthesis.

\noindent	\textbf{Human Motion Prediction.} Human motion prediction \cite{ma2022progressively,li2021multiscale, gu2022learning,yuan2020dlow, mao2021generating, gu2024learning,xu2022diverse, dang2022diverse,mao2021multi, xu2023joint,yu2023towards} has been actively studied in recent years due to the rapid advance of deep learning. It can be categorized into two directions: deterministic and stochastic. In deterministic scenarios, previous studies forecast one single future motion from a given observation sequence. They typically employ powerful network architectures to capture temporal dependencies from the history to regress the accurate future motion, such as graph convolutional networks \cite{zhong2022spatio,ma2022progressively,li2021multiscale} or Transformers \cite{mao2021multi, xu2023joint,yu2023towards}. Because human motion prediction is inherently ill-posed, later efforts switched to taking a stochastic approach  to this task by enabling diverse predictions. In the context of stochastic prediction, prior works \cite{yuan2020dlow, mao2021generating, gu2024learning,xu2022diverse, dang2022diverse} mostly exploit deep generative models to characterize the multimodal nature of human motion. Yuan et al. \cite{yuan2020dlow} designed learnable mapping functions to cover post-hoc sample diversity from a pre-trained generator. Wei et al. \cite{mao2021generating} proposed sequentially assembling body subsets for both controllable and diverse predictions. 

These works, however, do not reflect any semantic clues for prediction. Recently, Wei et al. \cite{mao2022weakly} introduced a new task, action-driven stochastic motion prediction and a resulting approach to it. The primary challenge in this task stems from the lack of transition data between two arbitrarily labeled action sequences in the available datasets. To address this, \cite{mao2022weakly} proposed paring one history with another future sequence of different action types as alternative history-future training data. It then simply \guPr{encourages} motion smoothness between such pairs for natural transition. Nevertheless, smoothness only does not ensure the motion validity, especially when the history and future sequences differ in orientation. As such, it can easily induce invalid motions within prediction, e.g., foot sliding, which lowers its applications.
In the context of the research goal, \cite{mao2022weakly} is the closest work to ours. However, our method regards the transition modeling as an in-betweening learning task to endow the transition with required leg dynamics, which fundamentally differs from \cite{mao2022weakly} and \guMR{yields} more cohesive and natural transitions.

\noindent	\guMR{\textbf{Human Motion Synthesis.} The synthesis of human motion focuses on the generation of history-free motions. Due to the recent advance of diffusion-based generative approaches \cite{karunratanakul2023guided, zhou2023unified, guo2022generating, alexanderson2023listen, yuan2023physdiff}, this field has gained huge progress especially regarding the motion realism. Furthermore, similar to the multimodality learning \cite{
yu2014click, zhang2021vector} in the domain of images, recent methods often condition multimodal input signals during synthesis, such as audio \cite{alexanderson2023listen} or text \cite{yuan2023physdiff, guo2022generating}. Alexanderson et al. \cite{alexanderson2023listen} modeled complex co-speech gesticulation or dancing motions which co-occurs with audio using diffusion models. Yuan et al. \cite{yuan2023physdiff} incorporated physical constraints as a refining process to mitigate the issues of motion floating and penetration. These methods, however, do not require learning transitions between multiple motion clips, which is primarily different from our focus.}

\noindent	\textbf{Human Motion In-betweening.} Different from motion prediction, human motion in-betweening aims to in-fill the missing intermediate frames given observed motion \guPr{clips} or pose cues. Trivial attempts include spline Bezier interpolation for keyframes, which is broadly adopted in the animation industry (e.g., Maya). However, it often entails tedious tuning efforts for better control \cite{qin2022motion}. Analogous to the prediction task, recent works \cite{kim2022conditional, qin2022motion, harvey2018recurrent, harvey2020robust} \guPr{have also} benefited from the development of deep neural networks. Harvey et al. \cite{harvey2018recurrent} proposed Recurrent Transition Networks (RTN), which is built upon Long-Short Term Memory (LSTM), to auto-regressively complete the frames within contexts. It was later improved in \cite{harvey2020robust} by applying time-to-arrival embedding for further transition smoothness. Duan et al. \cite{duan2022unified} \guPr{incorporated} a mixture embedding layer into the Transformer architecture to complete long-term missing frames in a non-autoregressive fashion. Qin et al. \cite{qin2022motion} formulated a two-stage Transformer-based framework for detail refinement. To further allow the in-betweening to respect semantic behavior, Kim et al. \cite{kim2022conditional} developed \cite{harvey2020robust} such that action type can \guPr{be} depicted during infilling. In this sense, \cite{kim2022conditional} is the most closely related method to our in-betweening learning model, particularly concerning conditioning on semantic labels, whereas the stark difference is that our AinB-VAE is a stochastic framework. It is note-worthy that stochastic in-betweening approaches have been studied sparsely to date, with the only exceptions being \cite{harvey2020robust, ren2023diverse}. Yet, these methods do not identify semantics. In contrast \guPr{to} the above techniques, our in-betweening model addresses both transition stochasticity and semantic clues, and is designed to tackle large orientation misalignment.

\noindent	\textbf{Motion Transition Learning.} It is still challenging for existing methods to synthesize long-term motion given a stream of action labels with natural transitions. Generally, prior perdition/synthesis models can be trained to compose multiple actions once the transition data for supervision is provided. Lee et al. \cite{lee2023multiact} formatted the data into (history, transition, future) such that the generator can learn with the ground-truth supervision triplet. However, \guPr{the} existing dataset (i.e., BABEL \cite{punnakkal2021babel}) only contains such transition data between very limited actions, and enriching the dataset to cover all possible transitions is less practical to realize. Consequently, these models exhibit poor generalization capacity. To nonetheless devise a robust transition learning policy, later efforts attempted to explore supervision-free methods. Athanasiou et al. \cite{athanasiou2022teach} trained the generative model by using the long-term sequences with multiple actions in BABEL and \guPr{employed a} trivial interpolation scheme (i.e., Slerp) to address the non-continuity. Mao et al. \cite{mao2022weakly} resorted to DCT smoothness prior for frequency-guided weak supervision. Li et al. \cite{li2023sequential} transferred the end pose of previous motion to the next pose for cohesive transition. Still, these methods do not ensure natural transitions when significant orientation changes exist, resulting in invalid leg dynamics. By contrast, our method is formulated to specifically model required leg movements with a dedicated orientation warping module for further realism.

%natural human motion can orient to any direction
%lowering naturalness

%In principle, given supervised data covering all possible action transitions, they could be trained to generate more complex motions. However, such data cannot practically be obtained. We overcome this by designing a weakly-supervised training strategy that lets us leverage limited, single-action sequences.

\section{Method}

\begin{figure*}[t]
\begin{center}
\includegraphics[width=0.9\linewidth]{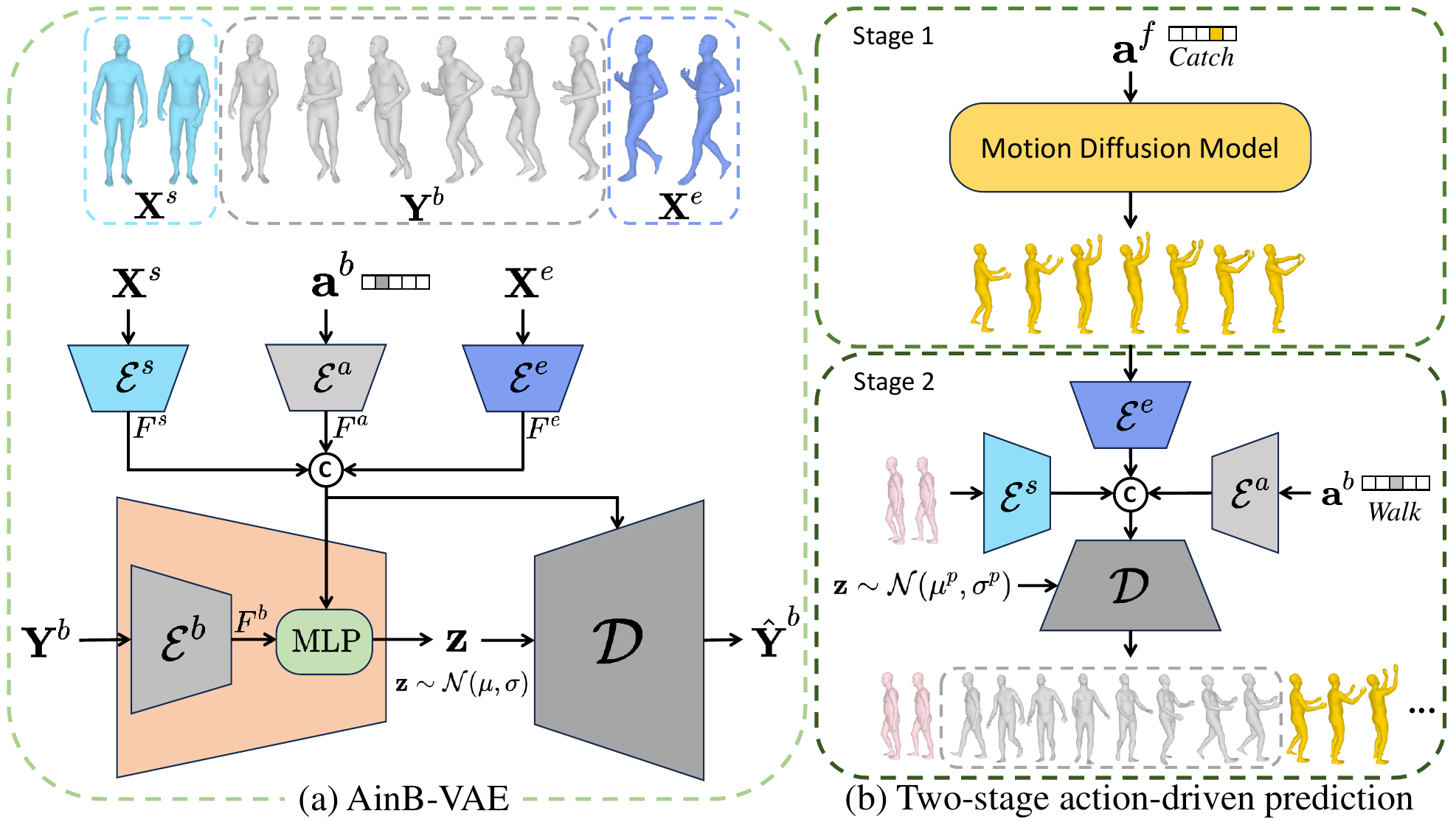}
\end{center}
\caption{Overview of our method. (a) depicts our transition generation framework$-$AinB-VAE. (b) gives the pipeline of our action-driven human motion prediction, which involves two stages. The two stages can be alternately performed to predict long-term motions with natural transitions in a recursive fashion.  }
\label{fig:overview}
\end{figure*}

We now introduce our method for action-driven
stochastic human motion prediction. Given a history sequence $\mathbf{X} = (\mathbf{x}_1, \mathbf{x}_2, \cdots, \mathbf{x}_{T_h})$ with all ${T_h}$ observed timesteps, we aim to forecast the future sequence $\mathbf{Y} = (\mathbf{y}_1, \mathbf{y}_2, \cdots, \mathbf{y}_{T_f})$, which \guPr{has ${T_f}$ timesteps in total}. Each frame in both sequences $\mathbf{x}_i, \mathbf{y}_i \in \mathbb{R}^{N}$ \guPr{has} $N$ dimensions. We follow the SMPL body model  \cite{loper2015smpl} for per-frame 3D mesh parameterization regarding the shape and pose. Importantly, the future sequence is also required to respect an additionally given one-hot encoded action label $\mathbf{a}^{f}$. As illustrated in Fig. \ref{fig:overview}(b), our method involves two stages in which we generate realistic transition and prediction results, respectively. As will be detailed below, our transition learning strategy requires no ground-truth transition annotated data and instead explores natural leg transition dynamics from individual-action motion data. 

\subsection{Orientation-aware action-conditioned transition generation}
\noindent	\textbf{Overview.} Human motion naturally involves the transition to bridge two motion actions. However, simply encouraging smoothness between two actions \cite{mao2022weakly} would induce unrealistic transitions which appear like the human body is forcibly ``dragged'' to stitch them with strong foot sliding. This issue is aggravated when the orientation misalignment increases. To address this issue, we argue that despite the lack of ground-truth labeled transition data, the generation of natural transition can be alternatively achieved by learning motion in-betweening from these individual actions. 
Therefore, we propose a conditional VAE (CVAE) framework to model motion in-betweening. Fig. \ref{fig:overview}(a) illustrates our model. Motivated by the empirical observation that valid human motion to account for orientation changes would inevitably involve leg dynamics, our CVAE is only modeled to characterize the in-betweening for some specific action types whose legs movements are sufficiently rich, such as ``\textit{Walk}'' and ``\textit{Run}''. \textit{In essence, our key insight is to endow  the action transition with required leg motions such that the movement appears natural when the human body turns \guPr{in} different directions.} As will be shown in our experiments, learning on these limited action categories suffices to generalize to other unseen action types for realistic in-betweening generation.
%Consequently, the transition can largely disobey the required human dynamics in switching body orientations. 

\noindent	\textbf{CVAE Modeling for Transition Generation.} 
Formally, given a one-hot encoded in-betweening action label $\mathbf{a}^b$ and a start-end sequence pair $(\mathbf{X}^s =  (\mathbf{x}^{s}_1, \mathbf{x}^{s}_2, \cdots, \mathbf{x}^{s}_{T_s}), \mathbf{X}^e =  (\mathbf{x}^{e}_1, \mathbf{x}^{e}_2, \cdots, \mathbf{x}^{e}_{T_e}))$ with, respectively, the length of $T_s$ and $T_e$, our CVAE formulates the conditional generative model $p_{\theta}(\mathbf{Y}^{b}|\mathbf{X}^{s}, \mathbf{X}^{e}, \mathbf{a}^b)$ for the in-betweening $\mathbf{Y}^{b} =  (\mathbf{y}^{b}_1, \mathbf{y}^{b}_2, \cdots, \mathbf{y}^{b}_{T_b})$ with $T_b$ frames, where $\theta$ denotes the model parameters. To characterize the transition stochasticity and multi-modality of the behavior in each action category, we follow the VAE learning policy \cite{kingma2013auto} by introducing the latent variable $\mathbf{z}$ and rewrite the conditional distribution as 
\begin{equation}
\label{eq:eq1}
p_{\theta}(\mathbf{Y}^{b}|\mathbf{X}^{c}, \mathbf{a}^b) = \int_{\mathbf{z}} p_{\theta}(\mathbf{Y}^{b}|\mathbf{X}^{c}, \mathbf{a}^b, \mathbf{z})p_{\theta}(\mathbf{z}|\mathbf{a}^b, \mathbf{X}^{c})d\mathbf{z},
\end{equation}
in which $\mathbf{X}^{c} = \{\mathbf{X}^{s}, \mathbf{X}^{e}\}$ represents the context information. $p_{\theta}(\mathbf{z}|\mathbf{a}^b,\mathbf{X}^{c})$ is a conditional Gaussian prior. \guMR{To address the intractability of Eq. \ref{eq:eq1}, we follow \cite{kingma2013auto} by optimizing the Evidence LOwer Bound (ELBO) as the CVAE training objective. This is realized by taking the logarithm of Eq. \ref{eq:eq1} and rewriting it to 
\begin{align}
\begin{split}
\label{eq:eq_extra}
\log{p_{\theta}(\mathbf{Y}^{b}|\mathbf{X}^{c}, \mathbf{a}^b)} = &\log \int_{\mathbf{z}}\frac{ q_{\phi}(\mathbf{z}|\mathbf{X}^{c}, \mathbf{Y}^{b}, \mathbf{a}^b)}{ q_{\phi}(\mathbf{z}|\mathbf{X}^{c}, \mathbf{Y}^{b}, \mathbf{a}^b)} \times \\
&p_{\theta}(\mathbf{Y}^{b}|\mathbf{X}^{c}, \mathbf{a}^b, \mathbf{z})p_{\theta}(\mathbf{z}|\mathbf{a}^b, \mathbf{X}^{c})d\mathbf{z},
\end{split}
\end{align}
where $q_{\phi}(\mathbf{z}|\mathbf{X}^{c}, \mathbf{Y}^{b}, \mathbf{a}^b)$ refers to an $\phi$-parameterized approximate posterior distribution. By further applying Jensen's inequality to Eq. \ref{eq:eq_extra}, the ELBO we need to  \textit{maximize} is given by}: 
\begin{equation}
\label{eq:eq2}
\begin{aligned}
\mathcal{L}_{ELBO} =& \mathbb{E}_{q_{\phi}(\mathbf{z}|\mathbf{X}^{c}, \mathbf{Y}^{b}, \mathbf{a}^b)}[\log p_{\theta}(\mathbf{Y}^{b} | \mathbf{z}, \mathbf{X}^{c}, \mathbf{a}^b)] \\
- &\mathrm{KL}( q_{\phi}(\mathbf{z}|\mathbf{X}^{c}, \mathbf{Y}^{b}, \mathbf{a}^b) || p_{\theta}(\mathbf{z}|\mathbf{a}^b,\mathbf{X}^{c})).
\end{aligned}
\end{equation}
In Eq. \ref{eq:eq2}, the $\mathrm{KL}(\cdot||\cdot)$ measures the Kullback–Leibler (KL) divergence between two distributions.  

Our Action-conditioned in-Betweening Variational Auto-encoder (AinB-VAE) learns to respect \guPr{the} action label for stochastic in-betweening generation. As shown in Fig. \ref{fig:overview}(a), AinB-VAE  by design consists of a context encoder pair, an action encoder, an in-betweening encoder, and an in-betweening decoder. We introduce in detail  AinB-VAE in the following \guPr{part} of this section.

\noindent	\textbf{AinB-VAE Context Encoding.} We start with introducing the context encoding in AinB-VAE. Since the context information contains both start and end sequences, we respectively design two context encoders, i.e., start and end \guPr{encoders}, for embedding learning. We utilize the Transformer architecture by enforcing multi-head self attention (MHSA) mechanism \cite{vaswani2017attention} for context encoding. We introduce a learnable embedding token $tok^{s}$ \cite{athanasiou2022teach, petrovich2022temos} as the prefix for the start motion $\mathbf{X}^s$ by injecting in $tok^s$ the temporal context dependencies. The periodic position encoding \cite{fan2022faceformer} is adopted to inform the Transformer with the timestep of each pose in context sequence. The start encoder $\mathcal{E}^s$ receives the $({tok^{s}, \mathbf{x}^s_1, \cdots, \mathbf{x}^s_{T_s}})$ to output the embedding $F^s \in \mathbb{R}^{D^s}$. Similarly, the end encoder $\mathcal{E}^e$ follows the identical encoding pipeline to the start encoder to produce $F^e \in \mathbb{R}^{D^e}$.

\noindent	\textbf{AinB-VAE Action Encoding.} We notice that the action encoding policy in ACTOR \cite{petrovich2021action} that employs action tokens reflects poor action faithfulness during generation. This is because AinB-VAE is required to respect three-fold conditions, i.e., start/end contexts and action label, which is inherently more complex than the case for ACTOR. Therefore, alike to ACTOR, we directly enforce a multilayer perceptron (MLP) to encode the in-betweening action $\mathbf{a}^b$, and then condition the resulting embedding in both encoding and decoding stages. Specifically, the action encoder $\mathcal{E}^a$ embeds the action representation $F^a$ with $D^a$ dimensions, following $F^a = \mathcal{E}^a(\mathbf{a}^b)$.

\noindent	\textbf{AinB-VAE In-betweening Encoding.} The overall In-betweening encoding policy follows that for the contexts. The only difference is that we additionally use the periodic causal mask \cite{fan2022faceformer} to better characterize long-term inter-frame dependencies, by biasing the attention which has closer period with higher weights. In particular, the output of In-betweening encoder $\mathcal{E}^b$ is obtained via $F^b = \mathcal{E}^b(\mathbf{Y}^b)$ with $D^b$ dimensions. Given the learned temporal feature triplet $(F^s, F^e, F^b)$ and the action code representation $F^{a}$, the encoding stage eventually employs an MLP to yield the Gaussian parameters ($\bm{\mu}, \bm{\sigma}$) from the concatenated features $(F^s \textcircled{\small{c}} F^e \textcircled{\small{c}} F^b \textcircled{\small{c}} F^{a})$. The latent variable $\mathbf{z}$ is then achieved using the reparameterization trick \cite{kingma2013auto} to enable the AinB-VAE training. 

\begin{figure}[t]
\begin{center}
\includegraphics[width=1.0\linewidth]{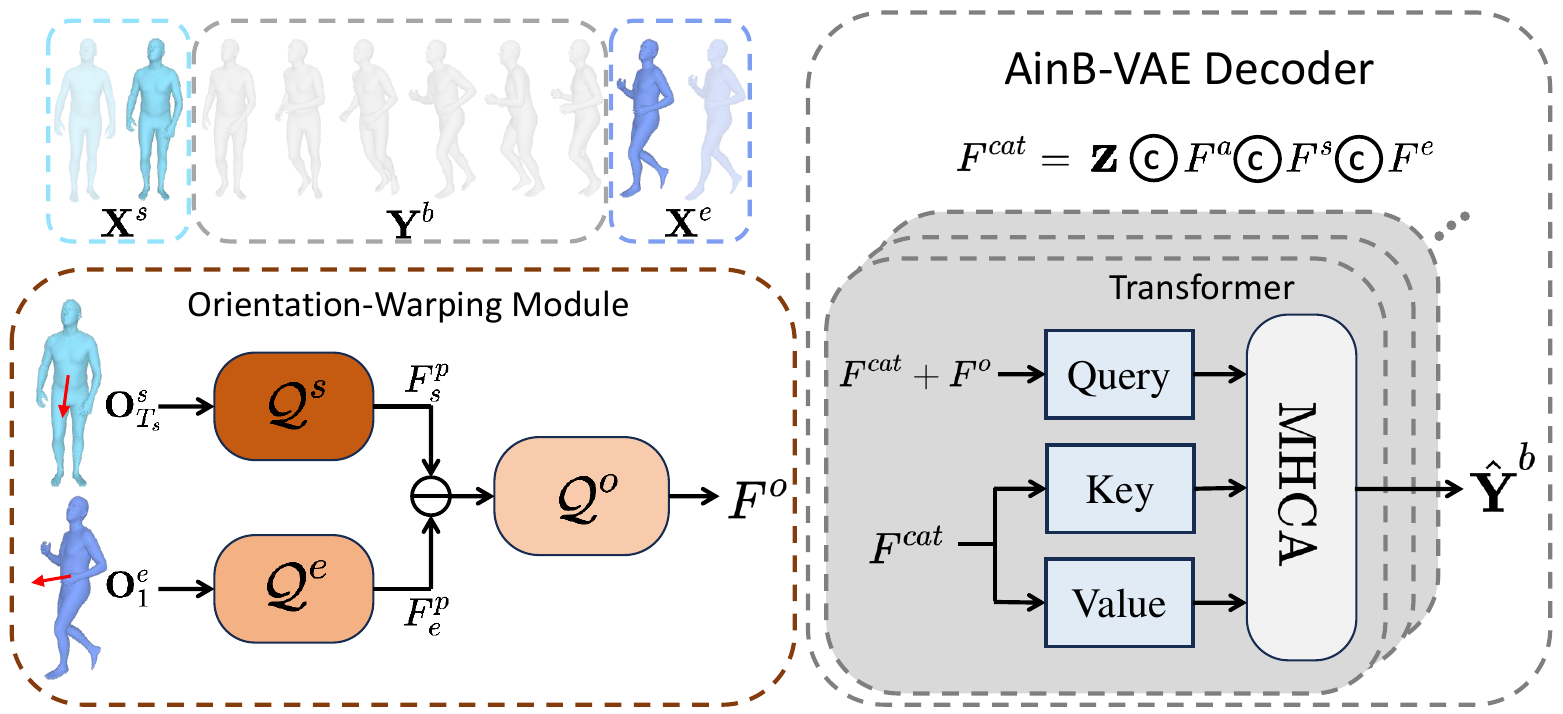}
\end{center}
\caption{Orientation-warping module (left) and the AinB-VAE Decoder (right). }
\label{fig:OWM}
\end{figure}

\noindent	\textbf{AinB-VAE Decoding.} 
\guPr{The} AinB-VAE decoding stage aims to reconstruct the in-betweening conditioned on the latent variable, motion context, and action label. A naive way would be decoding all the condition representations $F^{cat} = (\mathbf{z} \textcircled{\small{c}} F^{a} \textcircled{\small{c}} {F}^s \textcircled{\small{c}} {F}^e)$ with a MHSA for reconstruction. However, doing so cannot well model the bias in the context orientations since AinB-VAE targets intentionally on modeling in-betweenings whose context orientations differ noticeably. We thus design an orientation-warping module (OWM) to promote orientation-aware decoding, which attempts to inform the decoding stage with the orientation displacement. As depicted in Fig. \ref{fig:OWM}, 
\guPr{the} OWM consists of two parts: orientation feature extractor (OFE) and offset regressor (OR). We first map the orientation information into feature space with two OFEs $\mathcal{Q}^s, \mathcal{Q}^e$: 
\begin{equation}
\label{eq:eq3}
F^p_s = \mathcal{Q}^s(\mathbf{O}^s_{T_s}), F^p_e = \mathcal{Q}^e(\mathbf{O}^e_{1}),
\end{equation}
where $F^p_s$ and $F^p_e$ are with the same dimensionality of $D^s$. $\mathbf{O}^s_{T_s}$ and $\mathbf{O}^e_{1}$ are the global orientations of the last and the first frames in the start and end motions, respectively. The OR $ \mathcal{Q}^o$ then takes as input the displacement $F^p_e-F^p_s$ to regress the $D^o$-dimensional offset $F^o \in \mathbb{R}^{D^o}$: 
\begin{equation}
\label{eq:eq4}
F^o = \mathcal{Q}^o(F^p_e-F^p_s).
\end{equation}
\guMR{Here, all the mappings (i.e., $\mathcal{Q}^s$, $\mathcal{Q}^e$, and $\mathcal{Q}^o$) in the OFE and OR are implemented with MLPs.} In essence, $F^o$ reflects the required orientation change to form a realistic transition between the context motion pair.

Given that the condition representations $F^{cat}$ and the orientation bias $F^o$ naturally own multiple attributes, we replace the MHSA used in the encoding stage with Multi-Head Cross Attention (MHCA) to construct the decoder $\mathcal{D}$. Specifically, we feed $F^{cat}$ as key and value, and $F^o + F^{cat}$ as a query to provide orientation guide during decoding. \guMR{Importantly, the feature summing fuses two branches of embeddings with different attributes (i.e., the attributes from the condition and from the orientation guidance) to form an informative query. As such, the generated in-betweening is expected to yield further coherence.} The MHCA decoder eventually generates \guPr{a} sequence with $T_b$ poses to form the in-betweening reconstruction $\hat{\mathbf{Y}}^b$, following $\hat{\mathbf{Y}}^b = \mathcal{D}(\mathbf{X}^s, \mathbf{X}^e, \mathbf{a}^b, \mathbf{z})$.  

As explained in Eq. \ref{eq:eq2}, AinB-VAE is trained with two terms. The first term in Eq. \ref{eq:eq2} is simply measured with mean square error: $\mathcal{L}_{\mathrm{mse}} = ||\hat{\mathbf{Y}}^b - \mathbf{Y}^b||$. In designing the KL term, we enforce learnable prior by mapping the context and action encoding to Gaussian parameters ($\bm{\mu}^p, \bm{\sigma}^p$) with an MLP, which leads the Gaussian prior to be written \guPr{as} $p_{\theta}(\mathbf{z}|\mathbf{a}^b,\mathbf{X}^c) = \mathcal{N}(\bm{\mu}^p, \bm{\sigma}^p)$. The resulting KL terms can then be expressed as: $\mathcal{L}_{\mathrm{KL}} = \mathrm{KL}( \mathcal{N}(\bm{\mu}, \bm{\sigma}) ||  \mathcal{N}(\bm{\mu}^p, \bm{\sigma}^p))$. The final training objective for AinB-VAE to \textit{minimize} is given by:   
\begin{equation}
\label{eq:eq5}
\mathcal{L} = w_{\mathrm{mse}} * \mathcal{L}_{\mathrm{mse}} + w_{\mathrm{KL}} * \mathcal{L}_{\mathrm{KL}}, 
\end{equation}
where $w_{\mathrm{mse}}$ and $w_{\mathrm{KL}}$ are weighted to adjust the strength of each term.

\noindent	\textbf{In-betweening Sampling.} Once trained, AinB-VAE can produce plausible in-betweening based on the given conditions ${\mathbf{X}^c, \mathbf{a}^b}$ and the sampled latent variable $ \mathbf{z}$. Because AinB-VAE is designed to respect the action label, it provides two-fold transition multi-modalities in terms of inter- and intra-class diversity for the same context motion pair. We notice that AinB-VAE yields decent inter-class diversity via random sampling $\mathbf{z}$. Nevertheless, when provided with the same $\mathbf{a}^b$ and $\mathbf{X}^c$ that impose strong generation restrictions, the randomly sampled $\mathbf{z}$ can cause the produced in-betweenings to focus mostly on the major mode. To promote intra-class diversity, we adapt the diversity sampling technique \cite{yuan2020dlow} to our ACB generation setting by leveraging a motion sampler (MS) on the trained AinB-VAE. Specifically, \guPr{the} MS learns to map a sampled latent variable $\mathbf{z}$ to a set of latent variables $\{\mathbf{z}_1, \cdots, \mathbf{z}_l, \cdots, \mathbf{z}_L\}$ whose decoded in-betweening set  $\{\mathbf{Y}^b_1, \cdots, \mathbf{Y}^b_l, \cdots, \mathbf{Y}^b_L\}$ via AinB-VAE decoding is richly diversified. The mapping $r$, which is parameterized by $\eta$, performs linear transformation for $\mathbf{z}$ such that the mapped latent variables $r_{\eta}(\mathbf{z}_l|\mathbf{X}^c, \mathbf{a}^b)$ stay Gaussian distributed to circumvent domain shift. We give our sample loss in the following: 
\begin{equation}
\label{eq:eq6}
\begin{aligned}
\mathcal{L}_{samp} &= w_{\mathrm{div}} * \underset{i \neq j \in \lbrace1,\dots,L\rbrace}{\min}\norm{\mathbf{Y}^{b}_i-\mathbf{Y}^{b}_j}^2 \\
&+  w^{samp}_{\mathrm{KL}} * \sum_{l=1}^{L} \mathrm{KL}( r_{\eta}(\mathbf{z}_{l}|\mathbf{X}^c, \mathbf{a}^b) || p_{\theta}(\mathbf{z}|\mathbf{X}^c, \mathbf{a}^b)), 
\end{aligned}
\end{equation}
where ($ w_{\mathrm{div}},  w^{samp}_{\mathrm{KL}}$) are the weight pair. The first diversity term is designed to diversify intra-class transition variations. Note that our diversity term differs from \cite{yuan2020dlow} by penalizing the minimum pair-wise in-betweening distance to pursue duplication-aware diversification. The second KL term forces the transferred latent variables to remain the shape of Gaussian. 

\subsection{Action-driven stochastic human motion prediction}
\noindent	\textbf{Overview.} Since our goal is to provide a solution to stochastic action-driven human motion prediction, we hereby introduce our approach which leverages the trained AinB-VAE. In particular, we disentangle this task into two stages. We first synthesize the target motion $\mathbf{Y}$ following the given future action label $\mathbf{a}^f$. \guMR{Here, $\mathbf{a}^f$ is a one-hot encoded vector (i.e., $\mathbf{a}^f \in \mathbb{R}^{V^f}$) to facilitate an easy control of a total of $V^f$ types of possible future action.} We then regard the generated target motion $\mathbf{Y}$ and the given history motion $\mathbf{X}$ as the context motion pair, based on which our trained AinB-VAE decoder next produces the transition $\mathbf{X}^b$. In the following, we detail each stage.

\noindent	\textbf{Motion Diffusion for Synthesis.}
We employ the human motion diffusion model (MDM) \cite{tevet2023human} for action-conditioned motion synthesis. MDM follows the diffusion formulation \cite{sohl2015deep, song2020denoising} which models the stochastic diffusion process in thermodynamics for motion synthesis. It involves a diffusion process to add noise to the sample, and a reverse process for denoising. Specifically, in the diffusion process, an arbitrary training motion sample $\mathbf{Y}$ is progressively noised via the following Markov diffusion kernel:
\begin{equation}
\label{eq:eq7}
q(\mathbf{Y}_{t}|\mathbf{Y}_{t-1}) = \mathcal{N}(\sqrt{\alpha_t}\mathbf{Y}_{t-1}, (1-\alpha_t)I),
\end{equation}
where $t={1,\cdots,T}$ denotes an arbitrary timestep among a total of $T$ diffusion rounds and $\mathbf{Y}_0 = \mathbf{Y}$. \guMR{$\alpha_t \in (0,1)$ are scheduled constant hyperparameters in the diffusion formulation. In our implementation, we directly adopt the tuned scheduling configuration in MDM \cite{tevet2023human} and DDPM \cite{ho2020denoising} to ensure the best generation quality. Empirically, this is to circumvent that the forward and reverse processes have different functional forms that impose negative influence on the generation.} In practice, a sufficiently large $T$ results in the an approximation of diffused input $\mathbf{Y}$ to a random noise $\mathbf{s} \sim \mathcal{N}(0,I)$, i.e., $\mathbf{Y}_T \approx \mathbf{s}.$

The generation is via the reverse-diffusion process by denoising $\mathbf{s}$ back to a valid motion sample. Instead of modeling the inter-frame transition probability $p(\mathbf{Y}_{t-1}|\mathbf{Y}_{t})$, we leverage a $\psi$-parameterized generator $\mathcal{G}_{\psi}$ to directly recover the ``clean'' sample $\hat{\mathbf{Y}}_0$ from the noisy observation $\mathbf{Y}_t$ and with corresponding timestep information $t$:  $\hat{\mathbf{Y}}_0 = \mathcal{G}_{\psi}(\mathbf{Y}_t, t)$. For our conditional synthesis setting, we further condition the reverse-diffusion learning process with the future action label and rewrite the reverse-diffusion process as: $\hat{\mathbf{Y}}_0 = \mathcal{G}_{\psi}(\mathbf{Y}_t, t, \mathbf{a}^f)$. Due to the generator design, the training objective for MDM employs straightforward reconstruction supervision for the motion sample itself:
\begin{equation}
\label{eq:eq8}
\mathcal{L}_{mdm}=\mathbb{E}_{\mathbf{Y}_0, t \sim \{1, \cdots, T\}}||\mathcal{G}_{\psi}(\mathbf{Y}_t, t, \mathbf{a}^p) - \mathbf{Y}_0||^2.
\end{equation}
Note that we do not introduce any additional losses (e.g., geometric loss) as in \cite{tevet2023human} to be consistent with \cite{mao2022weakly}. The generator $\mathcal{G}$ adopts the MHSA architecture to characterize \guPr{the} temporal motion feature.

\noindent	\textbf{Two-stage Motion Prediction.} 
Up to this point, we have explained in detail our approach \guPr{to} transition learning, in-betweening sampler, and diffusion-based motion synthesis. 
We now explain how to utilize these components to fulfill the action-driven stochastic human motion prediction with a two-stage policy. Particularly, we first generate the target motion $\mathbf{Y}^{tar}$ with the action label $\mathbf{a}^f$ using the trained MDM generator. This is realized by feeding $\mathbf{s} \sim \mathcal{N}(0,I)$ as $\mathbf{Y}_T$ to the generator $\mathbf{Y}^{tar} = \mathcal{G}(\mathbf{Y}_T, T, \mathbf{a}^f)$. Then, we regard the given observed motion $\mathbf{X}$ as the start context  $\mathbf{X}^s$, and the synthesized $\mathbf{Y}^{tar}$ as the end context $\mathbf{X}^e$ to produce the transition with the trained AinB-VAE decoder by selecting one in-betweening category $\mathbf{a}^b$: $\mathbf{Y}^{b} =  \mathcal{D}(\mathbf{X}^s, \mathbf{X}^e, \mathbf{a}^b, \mathbf{z})$. The final motion prediction $\mathbf{Y}$ is achieved by $\mathbf{Y} = \mathbf{Y}^{b} \textcircled{\small{c}} \mathbf{Y}^{tar}$. As will be shown in our experiment, despite lacking the explicit ground-truth prediction supervision, it achieves comparable prediction accuracy with state-of-the-art methods. 

Importantly, our method presents three-fold diversity in forming the final prediction $\mathbf{Y}$ to provide a better modeling for stochasticity, regarding: (i) future motion diversity for $\mathbf{Y}^{tar}$ by re-sampling $\mathbf{s}$; (ii) \guMR{inter-class transition diversity, which refers to the diversity obtained by the  motion samples generated with different in-betweening action labels, for $\mathbf{Y}^{b}$ by altering the in-betweening label $\mathbf{a}^b$}; (iii) \guMR{intra-class transition diversity, which means the diversity measured by the  motion samples generated with the same in-betweening action label, for $\mathbf{Y}^{b} $ by diversity sampling $\mathbf{z}$ with in-betweening sampler.} Moreover, the two stages for prediction can be alternately performed to enable forecasting long-term sequence with a series of action labels.

%To realize this, we model the reverse transition probability $p_{\psi}(\mathbf{Y}_{t-1}|\mathbf{Y}_{t})$ with a ${\psi}$-parameterized deep neural network between adjacent timesteps, and iteratively performing the reverse-diffusion for $T$ times. For our conditional synthesis setting, we further condition the reverse-diffusion learning process with the future action label as $p_{\psi}(\mathbf{Y}_{t-1}|\mathbf{Y}_{t}, \mathbf{a}^p)$. The training of the diffusion model adopts the simple objective \cite{ramesh2022hierarchical} that directly optimizes the reconstruction $\mathbf{Y}_{0}$. Let the generator $\mathcal{G}$ denote the reverse modeling: $\mathcal{G}(\mathbf{Y}_t,t,c) = \mathbf{Y}_0$. The MDM training loss is given by

\section{Experiment}
In this section, we present extensive experimental results against prior methods to evaluate the effectiveness of our method. We also give detailed ablation studies to provide more understanding of our method. 

\begin{table*}[]
\caption{Quantitative evaluation of in-betweening quality on BABEL. FID (tr.) and FID (te.) refer to the FID scores obtained from the generation to train and test sets, respectively. We adapt RMI \cite{harvey2020robust} and MITT \cite{qin2022motion} to our task. }
\label{table1}
\centering
\scalebox{1.1}{
\begin{tabular}{cccccc:ccccc}
\toprule
               & \multicolumn{5}{c:}{$T_b$ = 40}            & \multicolumn{5}{c}{$T_b$ = 20}             \\
               & FID\_tr $\downarrow$ & FID\_te $\downarrow$ & AF $\uparrow$    & ADE $\downarrow$  & APD $\uparrow$  & FID\_tr $\downarrow$ & FID\_te $\downarrow$ & AF $\uparrow$   & ADE $\downarrow$  & APD $\uparrow$   \\ \hline
RMI   \cite{harvey2020robust}      & 37.09  & 30.15  & 1.51   & 0.979 & 0.079     & 37.87  & 33.43  & 0.91 & \textbf{0.599} & 0.051     \\
MITT   \cite{qin2022motion}      & 32.21  & 27.10  & 0.73  & 0.995 & -     & 40.62  & 36.64  & 0.66 & 0.658 & -      \\
CMIB    \cite{kim2022conditional}     & 61.07  & 53.60  & 1.96   & 1.215 & -     & 52.42  & 47.34  & 1.04 & 0.901 & -      \\ \hdashline
AinB-VAE       & \textbf{29.74}  & \textbf{23.53}  & 14.12 & \textbf{0.942} & 0.047 & \textbf{35.98}  & \textbf{31.62}  & 8.99 & 0.609 & 0.050 \\
AinB-VAE w. DS & 32.65  & 26.87  & \textbf{14.25} & 0.970 & \textbf{0.622} & 37.59  & 33.25  & \textbf{9.05} & 0.606 & \textbf{0.410} \\ \bottomrule
\end{tabular}}
\end{table*}

\noindent	\textbf{Dataset.} Following \cite{mao2022weakly}, we conduct the evaluation on three large action-labeled human motion datasets: BABEL \cite{punnakkal2021babel}, HumanAct12 \cite{guo2020action2motion}, and NTU RGB-D \cite{liu2019ntu, shahroudy2016ntu}. 

\textbf{BABEL} \cite{punnakkal2021babel} is a large-scale human motion dataset compromises nearly 40 hours \guPr{of} recording with frame-wise textual annotation for action. We down-sample all the sequences to 30 fps, and then remove  overly short ($<$ 1 second) ones. Because the sequences can contain multiple action \guPr{labels}, we split the long sequences into multiple short sub-sequences such that each individual sequence only follows one action label. Note that the transition-annotated \guPr{parts of the frames} are discarded and not involved during learning. We eventually have 20 actions. 10 frames are regarded as history in prediction.  %Note that although BABEL includes limited number of data with transition annotation between several actions, our method does not require such transition due to the poor generalization reported in \cite{mao2022weakly}.
 
\textbf{HumanAct12} \cite{guo2020action2motion} is 
adapted from PHSPD \cite{zou20203d} as a subset that contains 12 subjects in which 12 categories of actions with per-sequence annotation are provided. The sequences with \guPr{fewer} than 35 frames are removed, which results in 727 training and
197 testing sequences. Following \cite{mao2022weakly}, we use subjects P1 to P10 for training and P11, P12 for testing. 10 history frames are observed for prediction.

\textbf{NTU RGB-D} \cite{liu2019ntu, shahroudy2016ntu} contains over 100,000 motions with 120 classes whose pose annotations are from MS Kinect. This makes the data highly noisy and inaccurate. Some motions can be temporally inconsistent with severe jittering. We follow \cite{guo2020action2motion} by selecting 13 actions with 3900 motion clips, and the training/testing split policy of \cite{mao2022weakly}. 10 observed frames are provided for prediction.

In experimenting on these three datasets, we notice that the actions in HumanAct12 and NTU RGB-D are not quantitatively suited for learning decent leg movements since they mostly include the motions with upper-body variations. Therefore, AinB-VAE is only learned on the BABEL dataset. More specifically, in training the AinB-VAE, we select 4 actions: \textit{Walk, Jog, Run, and Step} from BABEL such that AinB-VAE can learn effectively from the sequences with rich leg dynamics. We then generalize the learned leg motions via AinB-VAE to HumanAct12 and NTU RGB-D to produce the transition. %As will be shown in our experiments, the transition generalization suffices to generate natural in-betweenings on unseen datasets. 

\noindent	\textbf{Implementation Details.} The training involves three models: AinB-VAE, MDM, and diversity sampler (DS). For AinB-VAE, we train it for 500 epochs with a learning rate of 0.001. The number of attention head is implemented with 4. ($w_{mse}$, $w_{\mathrm{KL}}$) are set to (100, 0.001). For MDM, we train it for 2000 epochs on BABEL and NTU RGB-D, and \guPr{for} 1000 epochs \guPr{on} HumanAct12, with \guPr{a} learning rate \guPr{of} 0.001. The total (reverse) diffusion is called for $T = 1000$ iterations. For DS, we set $(w_{div}, w_{\mathrm{KL}}^{samp})$ to (200,1) with \guPr{a} learning rate of 0.01 for 30 epochs. The AinB-VAE decoder is kept frozen during the sampler training. All the experiments are conducted on RTX4090.  

\subsection{Evaluation of in-betweening generation}
\label{sec:sec3.1}
Since the proposed AinB-VAE is, in essence, an in-betweening learning framework, we first evaluate the in-betweening generation performance. In particular, we compare AinB-VAE against three prior in-betweening learning methods: RMI \cite{harvey2020robust}, CMIB \cite{kim2022conditional}, and MITT \cite{qin2022motion} on the BABEL dataset. Specifically, for RMI \cite{harvey2020robust} and MITT \cite{qin2022motion}, we further adapt them to 
additionally take the action label as input to generate the in-betweening for fair comparisons. Specifically, we adopt the strategy in CMIB by first enforcing an MLP to obtain the embedding from the one-hot action vector. We retain the original architectures of RMI \cite{harvey2020robust} and MITT \cite{qin2022motion} for extracting features of observation (i.e., start and end) sequences, and add the action embedding to the observation features to regress the in-betweening, as suggested in  \cite{harvey2020robust, qin2022motion}. CMIB \cite{kim2022conditional} and MITT \cite{qin2022motion} are deterministic, whereas RMI \cite{harvey2020robust} and AinB-VAE allow stochastic in-betweening generation.

\noindent	\textbf{Evaluation Metrics.}  We follow previous motion synthesis \cite{guo2020action2motion}, prediction \cite{mao2022weakly}, and in-betweening \cite{harvey2020robust} learning methods by adopting the following metrics for evaluation: 
\begin{itemize}
    \item Frechet Inception Distance (FID). FID measures motion realism by computing the distance of two feature distributions between the generated motions and the real ones, following
    \begin{equation}
    \label{eq:eq9}
    \begin{split}
    FID = &||\mu_{gene} - \mu_{real}||^2 + \\
    &Tr(\Sigma_{gene} + \Sigma_{real} - 2(\Sigma_{gene}\Sigma_{real})^{1/2}),
    \end{split}
    \end{equation}
    where $Tr(\cdot)$ refers to the trace of a matrix.
    $\mu_{*} \in \mathbb{R}^{D^f}$ and $\Sigma_{*} \in \mathbb{R}^{D^f \times D^f}$ are the mean and covariance matrix that characterize the perception feature distribution with $D^f$ dimensions, which are acquired from a pre-trained action recognition model. We directly use the recognition model in \cite{mao2022weakly} for fair comparison. We report the FID of generation to train and test sets, respectively.

    \item Action Faithfulness (AF). We report the action recognition accuracy as the AF indicator for the generated in-betweening of all compared methods with the pre-trained action recognition model via \cite{mao2022weakly}. 

    \item  Average Displacement Error (ADE). We evaluate the ADE metric to examine the in-betweening accuracy of all methods. Given the generated in-betweening with $T_b$ frames, ADE is calculated \guPr{as} 
    \begin{equation}
    \label{eq:eq10}
    ADE = \frac{1}{T_b}\Sigma^{T_b}_{k=1}||y_k^b - \hat{y}_k^b||,
    \end{equation}
    which averages the $\mathcal{L}$2 displacement of predicted in-betweening and ground truth over all $T_b$ frames. For stochastic approaches (i.e., AinB-VAE and RMI \cite{harvey2020robust}), we follow \cite{yuan2020dlow,mao2022weakly} by reporting the minimum ADE. Given an in-betweening set $\{(\hat{\mathbf{Y^b}})^s\}_{i=1}^S$ with $S$ generated samples, we report $ADE = \underset{i = 1, \cdots, S}{min}\frac{1}{T_b}\Sigma^{T_b}_{k=1}||{{y_k^b}}^{(i)} - \hat{y}_k^b||_2$ as the prediction accuracy.

    \item  Average Pairwise Distance (APD). We assess the per-action in-betweening diversity with APD for stochastic approaches, following
    \begin{equation}
    \label{eq:eq10}
    APD = \frac{1}{S(S-1)}\sum_{i=1}^S \sum_{j=1, j\neq i}^S||\hat{(\mathbf{Y^b}})^i-\hat{(\mathbf{Y^b}})^j||.
    \end{equation}
    APD measures the average $\mathcal{L}$2 distance between all the generation pairs to investigate diversity. 
\end{itemize}

\begin{figure}[t]
\begin{center}
\includegraphics[width=0.9\linewidth]{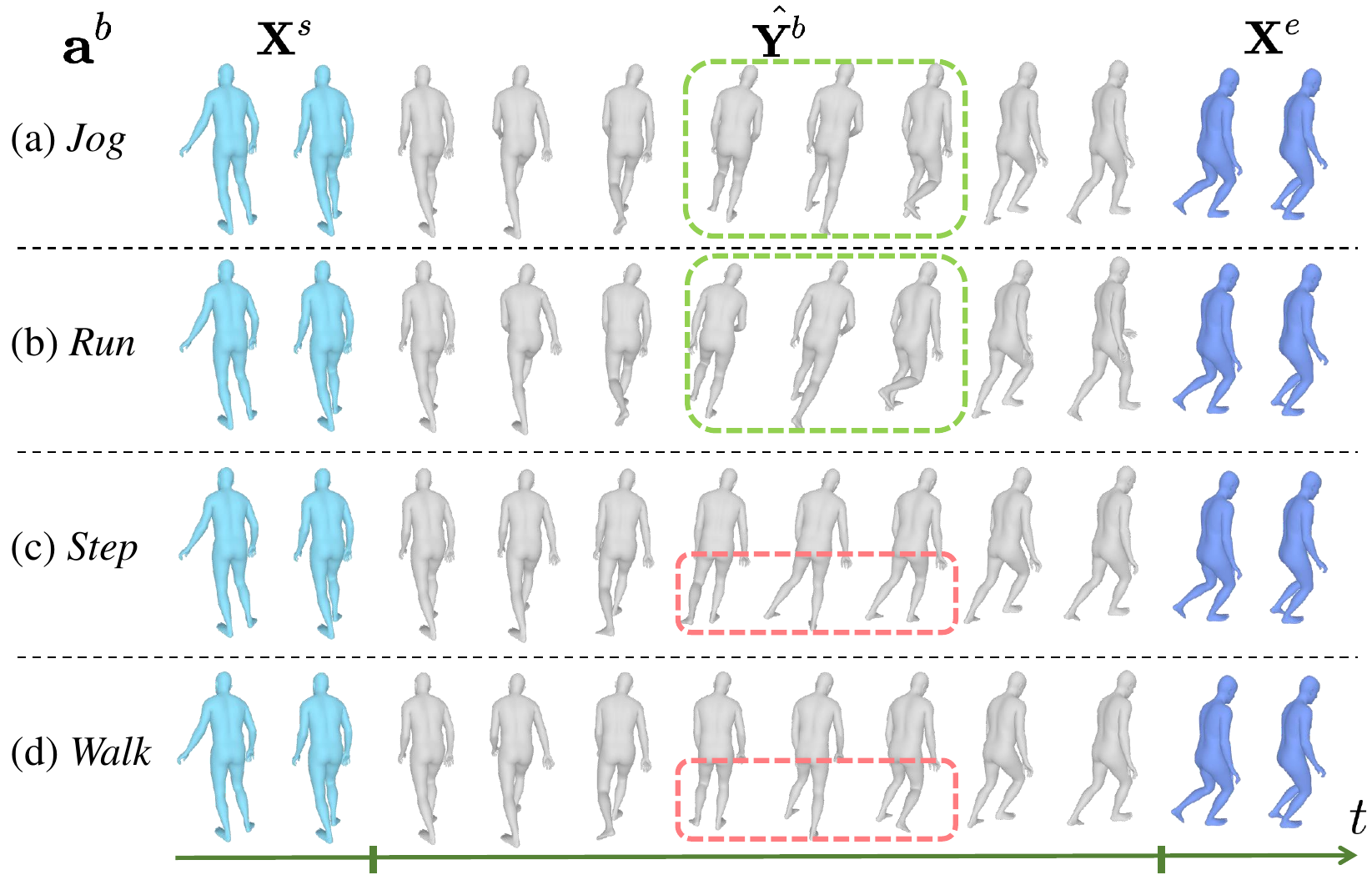}
\end{center}
\caption{Qualitative results of inter-action transition diversity given conditions $(\mathbf{X}_s, \mathbf{X}_e,\mathbf{a}^b)$ as input. }
\label{fig:inter_div}
\end{figure}

\begin{figure}[t]
\begin{center}
\includegraphics[width=0.9\linewidth]{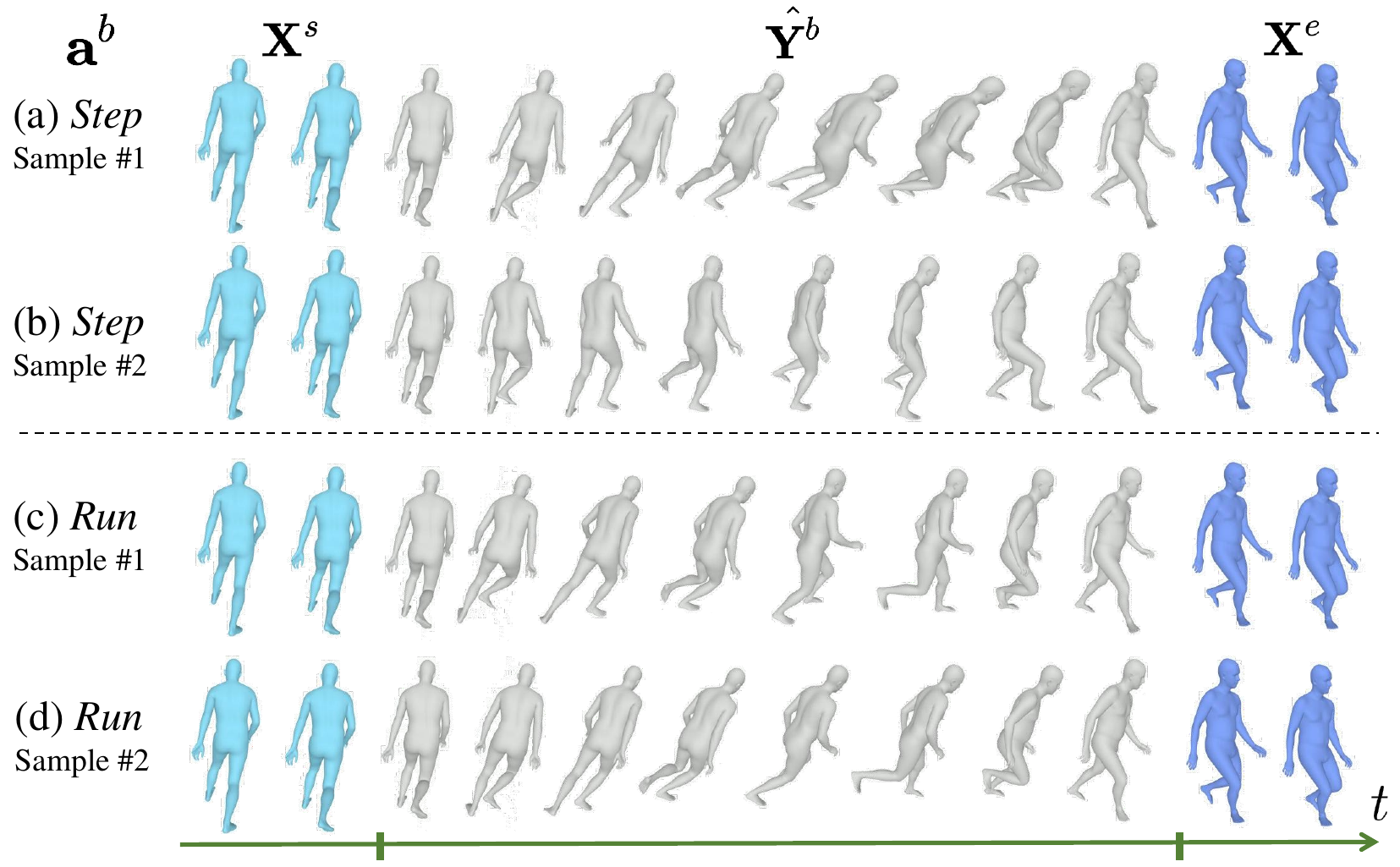}
\end{center}
\caption{Qualitative results of intra-action transition diversity by setting in-betweening action $\mathbf{a}^b$ to \textit{Step} and \textit{Run} as examples. All the results are obtained via diversity sampler. }
\label{fig:intra_div}
\end{figure}

\noindent	\textbf{Quantitative Results.} We here provide the quantitative evaluation for in-betweening generation performance. All the compared models are re-trained on the BABEL dataset with four action types (i.e., \textit{Walk, Jog, Run, and Step}) to ensure fair comparisons. The results are summarized in Tab. \ref{table1}. Given the same start $\mathbf{X}^s$ and end $\mathbf{X}^e$ motions with ($T_s, T_e$) set to (5,5), all methods generate $T_b$ frames of in-betweening. We prepare $T_b=40$  for long, and  $T_b=20$ for short transition settings.

It can be observed in Tab. \ref{table1} that our method outperforms the compared approaches in almost all evaluation metrics on two transition length settings. In particular, although RMI \cite{harvey2020robust} and MITT \cite{qin2022motion} achieve comparable generation quality to ours (3rd and 4th rows in Tab. \ref{table1}), the action faithfulness is less satisfactory. We assume that their frameworks are designed to specifically characterize natural in-betweenings, and reflecting accurate label information would require \guPr{a} further powerful conditioning strategy. \guPr{In addition}, for all methods, a longer in-betweening horizon results in better performance. This is because an increased frame capacity allows more flexibility for models to faithfully reflect the specified action category. Also, it can be confirmed from Tab. \ref{table1}(last column) that our sampler contributes to a significant diversity gain compared to the other stochastic approach RMI \cite{harvey2020robust}, with slightly sacrificing some FID scores.

\noindent	\textbf{Qualitative Results.} We next provide in Figs. \ref{fig:inter_div} and \ref{fig:intra_div} the qualitative results for better visual understanding. Fig. \ref{fig:inter_div} presents the inter-action label in-betweenings. We can see that for the same start and end condition motions, switching \guPr{the} transition label leads to action-faithful results. Specifically, even for  semantically similar action types, the produced transition still characterizes the subtle differences, such as the moving trend between \textit{Run} and  \textit{Jun} (green boxes in Fig. \ref{fig:inter_div}(a,b)). Also, it is interesting to point out that the \textit{Step} action label drives the foot motion to move sideways, which \textit{Walk} does not (magenta boxes
in Fig. \ref{fig:inter_div}(c,d)) achieve. We further visualize in Fig. \ref{fig:intra_div} the intra-class in-betweening results obtained via the diversity sampler. It can be observed that even under a triplet of condition signals, the sampler is still capable of diversifying the transitions by titling the body poses (e.g., Fig. \ref{fig:intra_div}(a,b)), yet respecting the given conditions. From the analysis above, we can thus confirm that our AinB-VAE yields two-fold transition multi-modalities in terms of iner- and intra- diversity with high motion realism. \textit{Please refer to the supplementary animation for a clear visual inspection, including the comparison against the compared methods.}

% \usepackage{multirow}
\begin{table}[]
\caption{Quantitative evaluation of action-driven motion prediction performance. All The results except for ours are reported directly from \cite{mao2022weakly}.}
\label{table2}
\centering
\scalebox{0.9}{
\begin{tabular}{ccccccc}
\toprule
                        &              & FID\_tr $\downarrow$       & FID\_te  $\downarrow$       & AF  $\uparrow$          & ADE    $\downarrow$       & APD   $\uparrow$        \\ \hline
\multirow{5}{*}{\rotatebox[origin=c]{90}{BABEL}}  & Act2Mot \cite{guo2020action2motion}     & 42.02          & 37.41           & 14.8          & 1.27          & 1.10          \\
                        & DLow \cite{yuan2020dlow}         & 27.99          & 24.18           & 12.7          & \textbf{1.19} & 0.9           \\
                        & ACTOR \cite{petrovich2021action}        & 29.34          & 30.31           & 40.9          & 2.29          & 2.71          \\
                        & WAT \cite{mao2022weakly} (RNN)    & 22.54          & 22.39           & 49.6          & 1.47          & 1.74          \\
                        & WAT \cite{mao2022weakly} (Trans.) & 20.02          & 19.41           & 39.5          & 1.40          & 1.82          \\ \hdashline
                        & Ours         & \textbf{16.39} & \textbf{19.12}  & \textbf{73.7} & 1.77          & \textbf{3.42} \\ \toprule
\multirow{5}{*}{\rotatebox[origin=c]{90}{HAct12}} & Act2Mot \cite{guo2020action2motion}      & 245.35         & 298.06          & 24.5          & 1.38          & 0.60          \\
                        & DLow \cite{yuan2020dlow}          & 254.72         & \textbf{143.71}          & 22.7          & 1.39          & 0.53          \\
                        & ACTOR \cite{petrovich2021action}        & 248.81         & 381.56          & 44.4          & 1.54          & 0.95          \\
                        & WAT \cite{mao2022weakly} (RNN)    & \textbf{129.95}         & 164.38          & 59.0          & \textbf{1.23}          & 0.96          \\
                        & WAT \cite{mao2022weakly} (Trans.) & 141.85         & 139.82          & 56.8          & 1.26          & 0.88          \\ \hdashline
                        & Ours         & 174.76         & 243.82          & \textbf{68.6} & 1.35          & \textbf{1.76} \\ \toprule
\multirow{5}{*}{\rotatebox[origin=c]{90}{NTU}}    & Act2Mot \cite{guo2020action2motion}      & 144.98         & 113.61          & 66.3          & \textbf{1.11} & 1.19          \\
                        & DLow \cite{yuan2020dlow}          & 151.11         & 157.54          & 70.6          & 1.20          & 1.21          \\
                        & ACTOR \cite{petrovich2021action}        & 355.69         & 193.58          & 66.3          & 1.49          & 2.07          \\
                        & WAT \cite{mao2022weakly} (RNN)    & \textbf{72.18} & \textbf{111.01} & \textbf{76.0} & 1.20          & 2.20          \\
                        & WAT \cite{mao2022weakly} (Trans.) & 83.14          & 114.62          & 71.3          & 1.23          & 2.19          \\ \hdashline
                        & Ours         & 374.73         & 530.09          & 63.9          & 1.41          & \textbf{2.73} \\ \bottomrule
\end{tabular}}
\end{table}

\begin{figure*}[t]
\begin{center}
\includegraphics[width=0.85\linewidth]{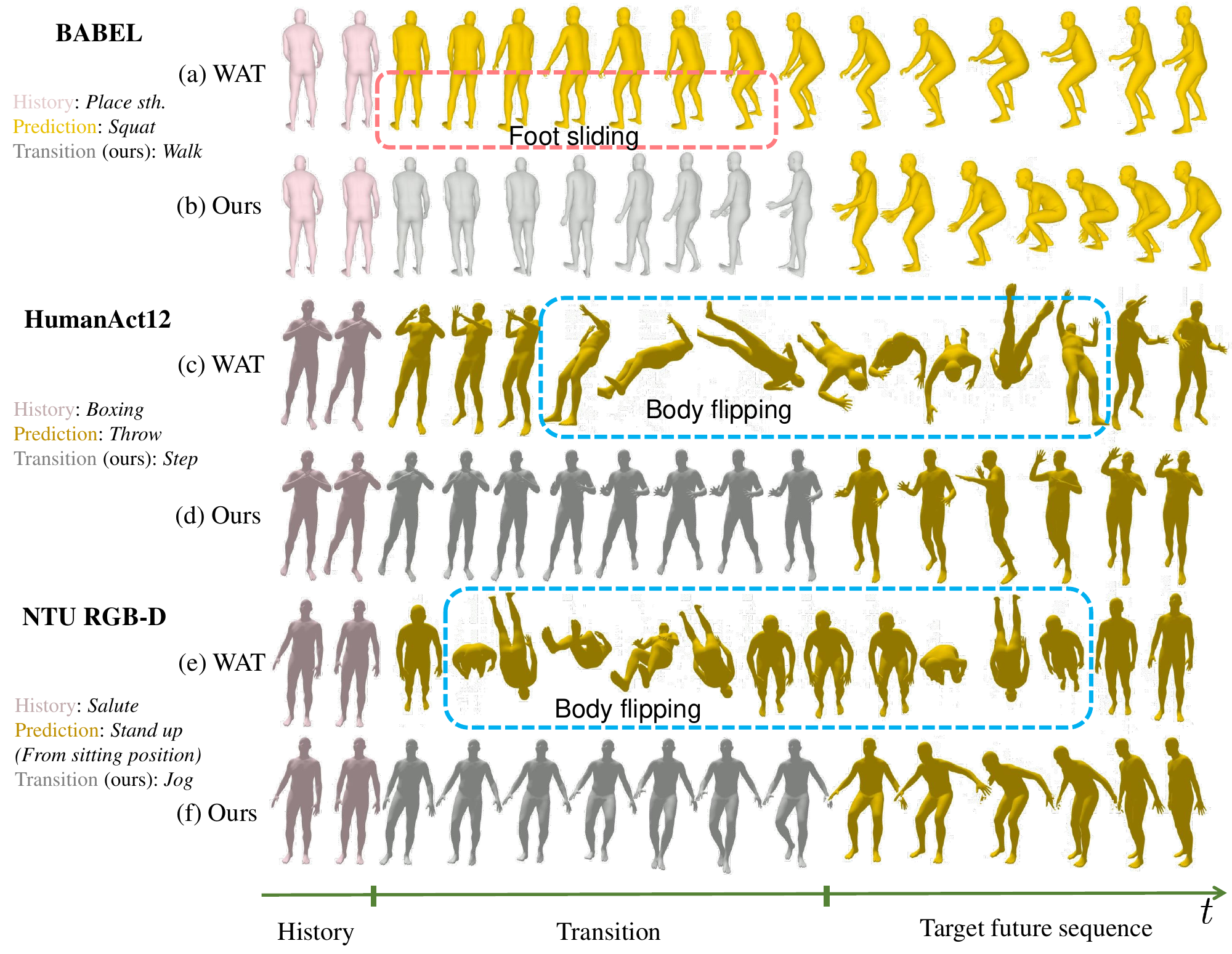}
\end{center}
\caption{Qualitative results of action-driven prediction against WAT \cite{mao2022weakly} on three datasets. We show in grey the produced transition with \guPr{the} learned AinB-VAE. 
The AinB-VAE is only trained on BABEL to generalize to HumanAct12 and NTU RGB-D for transition generation. See \guPr{the} supplementary animation for \guPr{a} better visual inspection.}
\label{fig:results_compare}
\end{figure*}

\subsection{Evaluation of action-driven human motion prediction} We here report the performance of action-driven human motion prediction. Specifically, we compare our two-stage prediction pipeline against prior arts: Action2Motion \cite{guo2020action2motion}, ACTOR \cite{petrovich2021action}, DLow \cite{yuan2020dlow}, and WAT \cite{mao2022weakly} on BABEL, HumanAct12, and NTU RGB-D. We follow the scheme in \cite{mao2022weakly} by predicting a stop sign to achieve variable length future motion synthesis. \textit{Note that we only train the AinB-VAE on BABEL and generalize it to HumanAct12 and NTU RGB-D for transition generation}. Similar to Sec. \ref{sec:sec3.1}, we adopt the metrics: FID, AF, ADE, and APD to be consistent with the compared approaches. The transition length $T_b$ is set to 40 to better reflect \guPr{the} in-betweening action type. Since our method is conceptually different from all the compared methods that either do not allow \cite{guo2020action2motion, petrovich2021action, yuan2020dlow} or involve very limited (i.e., $<$ 5 frames) \cite{mao2022weakly} transition length, during comparison, we directly assess the FID, AF, and ADE on the target sequence $\mathbf{Y}^{tar}$ instead of $\mathbf{Y}$ for consistent comparisons. The whole prediction sequence $\mathbf{Y}$ \guPr{that} involves both transition and target sequence is utilized for APD evaluation.

\noindent	\textbf{Quantitative Results.} 
The quantitative results for action-driven prediction \guPr{are} summarized in Tab. \ref{table2}. We can see that on the BABEL dataset, our method generally outperforms the prior methods in all metrics. However, our method performs less \guPr{satisfactorily} than existing methods on the NTU RGB-D dataset. \guMR{This is mainly because the motion data is more stable in BABEL, while in NTU RGB-D, it is  heavily polluted by noise during recording. The quantitative metric FID, which measures motion quality, computes the distance between the generation and the training/testing set. Since NTU RGB-D is highly noisy, a small FID would suggest that the generation may also reflect the noisy characteristics of the dataset. We expect this serves as the primary reason that our method performs well on BABEL, in which all the data remain stable, but achieves less desired quantitative results on NTU RGB-D. The qualitative results in the following also support our analysis.}

We next analyze the performance of prediction accuracy and diversity. Our method forecasts the whole future motion by 
sequentially generating the target and the transition sequences. Although we do not directly enforce supervision on each paired history and future sequences, as can be seen in Tab. \ref{table2}(6th column), our method achieves comparable or even \guPr{greater} (e.g., HumanAct12) prediction accuracy than those models trained with supervision. We assume the reason is that, relating a history with a randomly selected future sequence to form a history-future pair for supervision as in \cite{mao2022weakly} can hinder the model from learning the correctly annotated history-future dependency. We thus argue that our two-stage forecasting pipeline suffices to address the prediction accuracy for the task of action-driven motion prediction. \guPr{Regarding} diversity, it can be observed in Tab. \ref{table2}(last column) \guPr{that} our method consistently outperforms all other methods by a large margin. Similarly, the supervision in \cite{mao2022weakly} also sacrifices diversity for limited accuracy gain.

\noindent	\textbf{Qualitative Results.}
We provide in Fig. \ref{fig:results_compare} some qualitative results of action-driven motion prediction for visual inspection. For BABEL, we can see that our method and WAT both yield correct target action \guPr{categories} (Fig. \ref{fig:results_compare}(a,b)), yet WAT (a) triggers noticeable foot sliding (magenta box) to bridge the orientation gap between history and prediction. By contrast, our method achieves natural transition with valid leg movements for smooth connection (b). Let us now focus on HumanAct12 and NTU RGB-D, which both contain different degrees of noise. It can be observed in (c,e) that WAT 
induces highly unstable motions, such as body flipping, although they appear to follow the given action label. This is because the model learning is influenced too much by the noise within the training motion data. Our method, despite the fact that our AinB-VAE is trained on BABEL, generalizes well on unseen datasets to produce natural transitions (d,f). Besides, the generation stability significantly outperforms WAT, which further explains the underlying reason why the FID score is higher than \guPr{that for} WAT in Tab. \ref{table3}. The above analysis confirms that our method is capable of generating both more natural transition and accurate action-driven sequences to forecast coherent future motion, compared with the state of the art. We believe our strategy can also potentially benefit other relevant tasks, such as text-driven motion prediction.

\subsection{Ablation study} To gain more insights into our method, we perform the  following ablative experiments for \guPr{a} detailed assessment.

\noindent	\textbf{Design of OWM.}
OWM is designed to improve the in-betweening quality by imposing orientation guidance during decoding. To investigate this, we remove OWM and simply enforce MHSA for decoding to compare the performance. Moreover, 
OWM involves the orientation feature extractor (OFE) to handle orientation information in the feature space. We thus 
perform an additional ablation by removing the OFE and directly feeding the offset regresser with the orientation difference $\mathbf{O}^s_{T_s}-\mathbf{O}^e_{1}$ for decoding. The results are shown in Tab. \ref{table3}. It can be seen from Tab. \ref{table3} that both ablation scenarios lead to degraded in-betweening quality and a full OWM architecture contributes to the best performance. 
This indicates the effectiveness of introducing the OMW to \guPr{the} AinB-VAE decoding stage.

\begin{table}[]
\caption{Ablation study on the validity of OWM design on BABEL \cite{punnakkal2021babel}. OFE refers to the orientation feature extactor.}
\label{table3}
\centering
\begin{tabular}{cccccc}
\toprule
         & FID\_tr  $\downarrow$       & FID\_te   $\downarrow$      & AF $\uparrow$    & ADE $\downarrow$      & APD   $\uparrow$     \\ \hline
w. MHSA  & 32.34          & 26.16           & 13.78 & \textbf{0.942} & 0.046          \\
w.o. OFE  & 32.29          & 25.80          & \textbf{14.13} & \textbf{0.942} & \textbf{0.050} \\
AinB-VAE & \textbf{29.74} & \textbf{23.53} & 14.12 & \textbf{0.942} & 0.047          \\ \bottomrule
\end{tabular}
\end{table}

\begin{table}[]
\caption{Ablation study on the effectiveness of two-stage prediction pipeline on BABEL \cite{punnakkal2021babel}. PS denotes paired supervision.}
\label{table4}
\centering
\begin{tabular}{cccccc}
\toprule
         & FID\_tr  $\downarrow$       & FID\_te   $\downarrow$      & AF $\uparrow$    & ADE $\downarrow$      & APD   $\uparrow$     \\ \hline
w. PS  & 27.42          & 22.57          & 12.4 & \textbf{1.46} & 0.14 \\
w.o. PS (ours)  & \textbf{16.39} & \textbf{19.12} & \textbf{73.7} & 1.77 & \textbf{3.42}          \\ \bottomrule
\end{tabular}
\end{table}

\noindent	\textbf{History-Future Paired Supervision.} Our approach to action-driven prediction is designed to be without explicit history-future supervision. To verify the feasibility of this, we experiment by introducing such supervision for comparison. Specifically, we follow the training policy of \cite{mao2022weakly} by conditioning each future with a randomly selected history sequence with a different action label as paired supervision (PS) to re-train our motion diffusion model. As can be observed in Tab. \ref{table4}, despite the limited accuracy gain, training with PS would cause a noticeable loss in all other metrics, especially diversity (Tab. \ref{table4} last column).
Based on this observation, we can confirm that our two-stage prediction pipeline, which  does not even involve explicit history-future supervision, suffices to handle the task of action-driven prediction.

%Based on this observation, we design the instead of enforcing supervision to ensure better performance 

%we can confirm that action-driven prediction can be well handled even without explicit supervision.

\section{Conclusion}
We have presented a novel solution to the task of action-conditioned stochastic human motion prediction by focusing on the transition learning. It provides the insight that human motion transition to tackle orientation incoherence should naturally involve leg movements. Specifically, we divide this task into two folds by first generating a target motion with a specified action label, and then \guPr{producing} the in-betweening to specifically model the required leg dynamics. To model the stochasticity within transition, we formulate a VAE-based in-betweening framework and propose an orientation warping module to inform the decoding with orientation guidance. Our transition learning strategy does not demand the annotated transition data and is capable of reflecting the action type for transition. Moreover, our trained in-betweening model exhibits satisfactory generalization capacity. Experiments on three human motion datasets qualitatively and quantitatively demonstrate that our method achieves state-of-the-art performance for human motion in-betweening and prediction.

While our framework is effective in producing natural transitions, our in-betweening model does not infill variable length frames and follows a length-specific learning configuration. This could \guPr{potentially be} resolved by devising more powerful timestep embedding strategies. We would like to address this issue in the future.

% \fi

%\appendices

%\section{}


\ifCLASSOPTIONcaptionsoff
  \newpage
\fi

\bibliographystyle{abbrv}
\bibliography{reference}

\end{document}